\documentstyle[12pt,chicagob]{article}
\newcommand{\wt}{\mbox{\it wt}}
\newcommand{\fp}{$\mbox{FP}^{{\rm NP}[\log{n}]}$}
\newcommand{\fps}{\fp\ }
\newcommand{\fpsigma}{$\mbox{FP}^{\Sigma_2^P[\log{n}]}$}
\newcommand{\fpsigman}{$\mbox{FP}^{\Sigma_2^P[n]}$}
\newcommand{\fpsigmas}{\fpsigma\ }
\newcommand{\fpa}{$\mbox{FP}^{{\rm A}[\log{n}]}$}

\newcommand{\fpsigmaN}{$\mbox{FP}^{\Sigma_2^P[N\log{n}]}$}
\newcommand{\fpsigmaNs}{\fpsigmaN\ }
\newcommand{\fpN}{$\mbox{FP}^{{\rm NP}[N\log{n}]}$}
\newcommand{\fpn}{$\mbox{FP}^{{\rm NP}[n]}$}
\newcommand{\fpNs}{\fpN\ }

\newcommand{\fpsigmapar}{$\mbox{FP}^{\Sigma_2^P}_{||}$}
\newcommand{\fpsigmapars}{\fpsigmapar\ }
\newcommand{\fppar}{$\mbox{FP}^{{\rm NP}}_{||}$}
\newcommand{\fppars}{\fppar\ }

\newtheorem{theorem}{Theorem}[section]
\newtheorem{lemma}[theorem]{Lemma}
\newtheorem{definition}[theorem]{Definition}

\newtheorem{corollary}[theorem]{Corollary}

\def\squarebox#1{\hbox to #1{\hfill\vbox to #1{\vfill}}}
\newcommand{\qed}{\hspace*{\fill}
            \vbox{\hrule\hbox{\vrule\squarebox{.667em}\vrule}\hrule}\smallskip}
\newenvironment{proof}{\begin{trivlist}
\item[\hspace{\labelsep}{\bf\noindent Proof: }]
}{\qed\end{trivlist}}

\renewenvironment{proof}{\begin{trivlist}
\item[\hspace{\labelsep}{\bf\noindent Proof: }]
}{\qed\end{trivlist}}

\newtheorem{xmpl}[theorem]{Example}
\newenvironment{example}{\begin{xmpl}\rm}{\end{xmpl}}

\newtheorem{rmark}[theorem]{Remark}

\newcommand{\N}{\mbox{I$\!$N}}

\newcommand{\U}{{\cal U}}

\newcommand{\cS}{{\cal S}}

\newcommand{\Lan}{{\cal L}}
\newcommand{\sat}[1]{\|#1\|}
\newcommand{\zug}[1]{\langle #1  \rangle}

\newcommand{\stam}[1]{}
\newcommand{\bft}{{\bf true}}   
\newcommand{\bff}{{\bf false}}

\newcommand{\SH}{\mbox{{\it SH}}}
\newcommand{\BH}{\mbox{{\it BH}}}
\newcommand{\BT}{\mbox{{\it BT}}}
\newcommand{\BS}{\mbox{{\it BS}}}
\newcommand{\ST}{\mbox{{\it ST}}}

\newcommand{\bd}{\begin{definition}}
\newcommand{\ed}{\end{definition}}
\newcommand{\be}{\begin{enumerate}}
\newcommand{\bi}{\begin{itemize}}
\newcommand{\ee}{\end{enumerate}}
\newcommand{\ei}{\end{itemize}}

\newcommand{\cF}{{\cal F}}
\newcommand{\V}{{\cal V}}
\newcommand{\R}{{\cal R}}

\renewcommand{\phi}{\varphi}

\newcommand{\dr}{\mbox{{\em dr}}}
\newcommand{\db}{\mbox{{\em db}}}
\newcommand{\K}{{\cal K}}

\begin{document}

\begin{titlepage}
\title{Responsibility and Blame: A Structural-Model Approach}
\author{
Hana Chockler\\
School of Engineering and Computer Science\\
Hebrew University\\
Jerusalem 91904, Israel.\\ 
Email: hanac@cs.huji.ac.il
\and
Joseph Y. Halpern%
\thanks{Supported in part by NSF under grant
CTC-0208535 and by the DoD Multidisciplinary University Research
Initiative (MURI) program administered by ONR under
grant N00014-01-1-0795.}
\\
Department of Computer Science\\
Cornell University\\
Ithaca, NY 14853, U.S.A.\\
Email: halpern@cs.cornell.edu
}

\date{\today}

\maketitle
\thispagestyle{empty}

\begin{abstract}
Causality is typically treated an all-or-nothing concept; either $A$ is
a cause of $B$ or it is not.
We extend the definition of causality introduced by Halpern and Pearl
\citeyear{HP01b} to take into account the {\em degree of
responsibility\/} of $A$ for $B$.  For example, if someone wins an
election 11--0, then each person who votes for him is less responsiblefor the victory than if he had won 6--5.  We then define a notion of
{\em degree of blame}, which takes into account an agent's epistemic
state.  Roughly speaking, the degree of blame of $A$ for $B$ is the
expected degree of responsibility of $A$ for $B$,
taken over the epistemic state of an agent.
\end{abstract}
\end{titlepage}

\section{Introduction}

There have been many attempts to define {\em causality\/} going back to
Hume \citeyear{Hume39}, and continuing to the present (see, for example,
\cite{Collins03,pearl:2k} for some recent work). 
While many definitions of causality have been proposed, all of them 
treat causality is treated as an all-or-nothing concept.
That is, $A$ is either a cause of $B$ or it is
not.  As a consequence, thinking only in terms of causality does not at
times allow us to make distinctions that we may want to make.  For
example, suppose that Mr.~B wins an election against Mr.~G by a vote
of 11--0. 
Each of the people who voted for Mr.~B
is a cause of him winning.
However, it seems that their degree of responsibility should not be as
great as in the case when Mr.~B wins 6--5.  

In this paper, we present a definition of responsibility that takes this
distinction into account.   The definition is an extension of a
definition of causality
introduced by Halpern and
Pearl~\citeyear{HP01b}.   Like many other definitions of causality going back
to Hume \citeyear{Hume39}, this definition is based on counterfactual
dependence.  Roughly speaking, $A$ is a cause of $B$ if, had $A$ not
happened (this is the counterfactual condition, since $A$ did in fact
happen) then $B$ would not have happened.  As is well known, this naive
definition does not capture all the subtleties involved with causality.
(If it did, there would be far fewer papers in the philosophy
literature!)  
In the case of the 6--5 vote, it is clear that, according to this
definition, each of the voters for Mr.~B is a cause of him winning,
since if they had voted against Mr.~B, he would have lost.  On the
other hand, in the case of the 11-0 vote, there are no causes according
to the naive counterfactual definition.  
A change of one vote does not makes no difference.
Indeed, in this case, we do say in natural language
that the cause is somewhat ``diffuse''.

While in this case the standard counterfactual definition may not seem
quite so problematic, the following example (taken from \cite{Hall98}) 
shows that things can be even more subtle.  Suppose that
Suzy and Billy both pick up rocks and throw them at  a bottle.
Suzy's rock gets there first, shattering the
bottle.  Since both throws are perfectly accurate, Billy's would have
shattered the bottle had Suzy not thrown.
Thus, according to the naive counterfactual 
definition, Suzy's throw is not a cause of the bottle shattering.
This certainly seems counter to intuition.

Both problems are dealt with the same way in \cite{HP01b}. Roughly
speaking, the idea is that $A$ is a cause of $B$ if $B$ counterfactually
depends on $C$ {\em under 
some contingency}.  For example, voter 1 is a cause of Mr.~B winning
even if the vote is 11--0 because, under the contingency that 5 of the
other voters had voted for Mr.~G instead, voter 1's vote would have
become critical; if he had then changed his vote, Mr.~B would not have
won.  Similarly, Suzy's throw is the cause of the bottle
shattering because the bottle shattering counterfactually depends on
Suzy's throw, under the contingency that Billy doesn't throw.
(There are further subtleties in the definition that guarantee that, if
things are modeled appropriately, Billy's throw is not a cause.  These
are discussed in Section~\ref{sec:definitions}.)

It is precisely this consideration of contingencies that lets us define
degree of responsibility.  We take the degree of responsibility of $A$
for $B$ to be $1/(N+1)$, where $N$ is the minimal number of changes that
have to be made to obtain a contingency where $B$ counterfactually
depends on $A$.   
(If $A$ is not a cause of $B$, then the degree of responsibility is 0.)
In particular, this means that in the case of the 11--0 vote, the degree
of responsibility of any voter for the victory is $1/6$, since 5 changes
have to be made before a vote is critical.  If the vote were
1001--0, the degree of responsibility of any voter would be $1/501$.  On
the other hand, if the vote is 5--4, then the degree of responsibility
of each voter for Mr.~B for Mr.~B's victory is 1; each voter is
critical.  As we would expect, those voters who voted for Mr.~G have
degree of responsibility 0 for Mr.~B's victory, since they are not
causes of the victory.  Finally, in the case of Suzy and Billy, even
though Suzy is the only cause of the bottle shattering, Suzy's degree of
responsibility is $1/2$, while Billy's is 0.  Thus, the degree of
responsibility measures to some extent whether or not there are other
potential causes.

When
determining responsibility, it is assumed that everything relevant about
the facts of the world and how the world works (which we characterize in
terms of what are called {\em structural equations\/})
is known.  For example,
when saying that voter 1 has degree of responsibility $1/6$ for Mr.~B's
win when the vote is 11--0, we assume that the vote and the
procedure for determining a winner (majority wins) is known.  There is
no uncertainty about this.  Just as with causality, there is no
difficulty in talking about the probability that 
someone has a certain degree of responsibility by putting a
probability distribution on the way 
the world could be and how it works.  But this misses out on important
component of determining what we call here {\em blame}:~the epistemic
state.  Consider a doctor who treats a patient with a particular drug
resulting in the patient's death.  The doctor's treatment is a cause of
the patient's death; indeed, the doctor may well bear degree of
responsibility 1 for the death.  However, if the doctor had no idea that
the treatment had adverse side effects for people with  high
blood pressure, he should perhaps not be held to blame for the death.
Actually, in legal arguments, it may not be so relevant what the doctor
actually did or did not know, but what he {\em should have known}.
Thus, rather 
than considering the doctor's actual epistemic state, it may be more
important 
to consider what his epistemic state should have been.  But, in any
case, if we are trying to determine whether the doctor is to blame for
the patient's death, we must take into account the doctor's epistemic
state. 

We present a definition of blame that considers whether agent $a$
performing action $b$ is to blame for an outcome $\phi$.  The definition
is relative to an epistemic state for $a$, which is taken, roughly
speaking, to be a set of situations before action $b$ is performed,
together with a probability on them.  The degree of blame is then
essentially 
the expected degree of responsibility of action $b$ for $\phi$
(except that we ignore situations where $\phi$ was already true or 
$b$ was already performed).
To understand the difference between responsibility and blame, suppose
that there is a firing squad consisting of ten excellent marksmen.  Only
one of them has live bullets in his rifle; the rest have blanks.  
The marksmen do not know which of them has the live bullets.  The
marksmen shoot at the prisoner and he dies.  The only marksman that is
the cause of the prisoner's death is the one with the live bullets.
That marksman has degree of responsibility 1 for the death; all the rest
have degree of responsibility 0.  However, each of the marksmen has
degree of blame $1/10$.%
\footnote{We thank Tim Williamson for this example.}

While we believe that our definitions of responsibility and blame are
reasonable, they certainly do not capture all the connotations of these
words as used in the literature. 
In the philosophy literature, papers on responsibility typically are
concerned with {\em moral responsibility} (see, for example,
\cite{Zimmerman88}).  Our definitions, 
by design, do not take into account intentions or possible 
alternative actions,
both of which seem necessary in dealing with moral issues.  For example,
there is no question that Truman was in part responsible and to blame
for the deaths resulting from dropping the atom bombs on Hiroshima and
Nagasaki.  However, to decide whether this is a morally reprehensible
act, it is also  necessary to consider the alternative actions he could
have performed, and their possible outcomes.  While our
definitions do not directly address these moral issues, we believe that
they may be helpful in elucidating them.  
Shafer \citeyear{Shafer01} discusses a notion of responsibility that
seems somewhat in the spirit of our notion of blame, especially in that
he views responsibility as being based (in part) on causality. 
However, he does not give a formal definition of responsibility, so it
is hard to compare our definitions to his.  However, there are some
significant technical differences between his notion of causality 
(discussed in \cite{Shafer96}) and
that on which our notions are based.  We suspect that any notion of
responsibility or blame that he would define would be different from
ours. We return to these issues in Section~\ref{sec:discussion}.

The rest of this paper is organized as follows.  In
Section~\ref{sec:definitions}
we review the basic definitions of causal
models based on structural equations, which are the basis for our 
definitions of responsibility and blame.
In Section~\ref{sec:resp}, we review the definition of causality from
\cite{HP01b}, and show how it can be modified to give a definition of
responsibility.  
We show that the definition of responsibility gives
reasonable answer in a number of cases, and briefly discuss how it 
can be used in program verification (see \cite{CHK}).  
In Section~\ref{sec:blame}, we give our definition of blame.  In
Section~\ref{sec:complexity}, we discuss the complexity of computing
responsibility and blame.  
We conclude in Section~\ref{sec:discussion}
with some discussion of responsibility and blame.

\section{Causal Models: A Review}\label{sec:definitions}

In this section, we review the details of the definitions of causal models
from \cite{HP01b}.  
This section is essentially identical to the corresponding section in
\cite{CHK}; the material is largely taken from \cite{HP01b}.

A {\em signature\/} is a tuple $\cS = \zug{\U,\V,\R}$, 
where $\U$ is a finite set
of {\em exogenous\/} variables, $\V$ is a set of {\em endogenous\/}
variables,  
and $\R$ associates with every variable  
$Y \in \U \cup \V$ a nonempty set $\R(Y)$ of possible values for $Y$.
Intuitively, the  exogenous variables are ones whose values are
determined by factors outside the model, while the endogenous variables
are ones whose values are ultimately determined by the exogenous
variables.
A {\em causal model\/} over signature $\cS$ is a tuple
$M = \zug{\cS,\cF}$, where $\cF$ associates with every endogenous variable
$X \in \V$ a function $F_X$ such that 
$F_X: (\times_{U \in \U} \R(U) \times (\times_{Y \in \V \setminus \{ X \}}
\R(Y))) \rightarrow \R(X)$. That is, $F_X$ describes how the value of the
endogenous variable $X$ is determined by
the values of all other variables in $\U \cup \V$. 
If $\R(Y)$ contains only two values for each $Y \in \U \cup \V$, then 
we say that $M$ is a 
{\em binary causal model}.

We can describe (some salient features of) a causal model $M$ using a
{\em causal network}.  
This is a graph
with nodes corresponding to the random variables in $\V$ and an edge
from a node labeled $X$ to one labeled $Y$ if $F_Y$ depends on the value
of $X$.
Intuitively, variables can have a causal effect only on their
descendants in the causal network; if $Y$ is not a descendant of $X$,
then a change in the value of $X$ has no affect on the value of $Y$.
For ease of exposition, 
we restrict attention to what are called {\em
recursive\/} models. These are ones whose associated causal network
is a directed acyclic graph (that
is, a graph that has no  cycle of edges).
Actually, it suffices for our purposes that, for each setting $\vec{u}$
for the variables in $\U$, there is no cycle among the edges of causal
network.   
We call a setting $\vec{u}$ for the variables in $\U$ a {\em context}.
It should be clear that if $M$ is a recursive causal model,
then there is always a
unique solution to the equations in $M$, given a context.

The equations determined by $\{F_X: X \in \V\}$ can be thought of as
representing  processes (or mechanisms) by which values are assigned to
variables.  For example, if $F_X(Y,Z,U) = Y+U$ (which we usually write 
as $X=Y+U$), then if $Y = 3$ and $U = 2$, then $X=5$,
regardless of how $Z$ is set.
This equation also gives counterfactual information. It says that,
in the context $U = 4$, if $Y$ were $4$, then $X$ would
be $u+4$, regardless of what value 
$X$, $Y$, and $Z$
actually take in the real world. 

\begin{figure}[htb]
\input{psfig}
\centerline{\psfig{figure=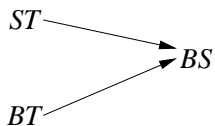}}
\caption{A naive model for the rock-throwing example.}\label{fig0}
\end{figure}

While the equations for a given problem are typically obvious, the
choice of variables may not be.  For example, consider the rock-throwing
example from the introduction.  In this case, a naive model might have
an exogenous variable $U$ that encapsulates whatever background factors
cause Suzy and Billy to decide to throw the rock
(the details of $U$
do not matter, since we are interested only in the context where $U$'s
value is such that both Suzy and Billy throw), a variable $\ST$ for Suzy
throws ($\ST = 1$ if Suzy throws, and $ST = 0$ if she doesn't), a
variable $\BT$ for Billy throws, and a variable $\BS$ for bottle shatters.
In the naive model, whose graph is given in Figure~\ref{fig0}, 
$BS$ is 1 if one of $\ST$ and $\BT$ is 1.
(Note that the graph omits the exogenous variable $U$, since it plays
no role.  In the graph, there is an arrow
from variable $X$ to variable $Y$ if the value of $Y$ depends on the
value of $X$.)

This causal model does not distinguish between Suzy and Billy's rocks 
hitting the bottle simultaneously and Suzy's rock hitting first.
A more sophisticated model might also include variables $\SH$ and $\BH$,
for Suzy's rock hits the bottle and Billy's rock hits the bottle.  
Clearly $\BS$ is 1 iff one of $\SH$ and $\BH$ is 1.  However, now,
$\SH$ is 1 if $\ST$ is 1, and $\BH = 1$ if $\BT = 1$ and $\SH = 0$.
Thus, Billy's
throw hits if Billy throws {\em and\/} Suzy's rock doesn't hit.
This model is described by the following graph,
where we implicitly assume a context where Suzy throws first, so there
is an edge from $\SH$ to $\BH$, but not one in the other direction.
\begin{figure}[htb]
\input{psfig}
\centerline{\psfig{figure=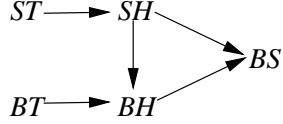}}
\caption{A better model for the rock-throwing example.}\label{fig1}
\end{figure}

Given a causal model $M = (\cS,\cF)$, a (possibly
empty)  vector
$\vec{X}$ of variables in $\V$, and vectors $\vec{x}$ and
$\vec{u}$ of values for the variables in
$\vec{X}$ and $\U$, respectively, we can define a new causal model
denoted
$M_{\vec{X} \gets \vec{x}}$ over the signature $\cS_{\vec{X}}
= (\U, \V - \vec{X}, \R|_{\V - \vec{X}})$.
Formally, $M_{\vec{X} \gets \vec{x}} = (\cS_{\vec{X}},
\cF^{\vec{X} \gets \vec{x}})$,
where $F_Y^{\vec{X} \gets \vec{x}}$ is obtained from $F_Y$
by setting the values of the
variables in $\vec{X}$ to $\vec{x}$.
Intuitively, this is the causal model that results when the variables in
$\vec{X}$ are set to $\vec{x}$ by some external action
that affects only the variables in $\vec{X}$;
we do not model the action or its causes explicitly.
For example, if $M$ is the more sophisticated model for the
rock-throwing example, 
then $M_{{\it ST} \gets 0}$ is the model where Suzy doesn't throw.

Given a signature $\cS = (\U,\V,\R)$, a formula of the form $X = x$, for
$X \in \V$ and $x \in \R(X)$, is called a {\em primitive event}.   A {\em
basic causal formula\/} 
has the form
$[Y_1 \gets y_1, \ldots, Y_k \gets y_k] \phi$,
where 
\begin{itemize}
\item 
$\phi$ is a Boolean
combination of primitive events;
\item $Y_1,\ldots, Y_k$ are distinct variables in $\V$;  and
\item 
$y_i \in \R(Y_i)$.
\end{itemize}
Such a formula is abbreviated as $[\vec{Y} \gets \vec{y}]\phi$.
The special
case where $k=0$
is abbreviated as $\phi$.
Intuitively, $[Y_1 \gets y_1, \ldots, Y_k \gets y_k] \phi$ says that
$\phi$ holds in the counterfactual world that would arise if
$Y_i$ is set to $y_i$, $i = 1,\ldots,k$.
A {\em causal formula\/} is a Boolean combination of basic causal
formulas.

A causal formula $\phi$ is true or false in a causal model, given a
context.
We write $(M,\vec{u}) \models \phi$ if
$\phi$ is true in
causal model $M$ given context $\vec{u}$.
$(M,\vec{u}) \models  [\vec{Y} \gets \vec{y}](X = x)$ if 
the variable $X$ has value $x$ 
in the unique (since we are dealing with recursive models) solution
to the equations in
$M_{\vec{Y} \gets \vec{y}}$ in context $\vec{u}$ (that is, the
unique vector
of values for the exogenous variables that simultaneously satisfies all
equations $F^{\vec{Y} \gets \vec{y}}_Z$, $Z \in \V - \vec{Y}$,
with the variables in $\U$ set to $\vec{u}$).
We extend the definition to arbitrary causal formulas
in the obvious way.

\stam{
We model {\em epistemic state\/} by a pair $\zug{U,D}$, where
$U$ is a set of contexts $U = \{ \vec{u_1}, \ldots, \vec{u_n} \}$ and 
$D$ is a probability distribution defined over $U$. For a subset of variables
$\vec{X}$ and an assignment $\vec{x}$ to $\vec{X}$, 
the {\em epistemic state of $(\vec{X}=\vec{x})$\/}
contains contexts where $\vec{X}$ is not set to $\vec{x}$. In other words, 
the epistemic state of $(\vec{X}=\vec{x})$ is the set of all possible contexts 
{\em before\/} $\vec{X}$ is set to $\vec{x}$.
In the example described in the introduction, 
a doctor's epistemic state would be the set of possible contexts
(or the contexts the doctor considers possible) before he administered 
the drug. The distinction of what the doctor knew versus what the doctor
should have known translates to the question of whether the current
context has to have a non-negative probability in the epistemic state.
The positive answer postulates that a doctor should have been aware to the
real situation before administering the drug, although he might have
considered other possible scenarios. That is, the doctor should have taken
into account the possibility that the patient has high blood pressure.
On the other hand, if we allow epistemic state to be based only on
the subjective knowledge, it is possible that the context where the
patient has a high blood pressure gets probability $0$ in the distribution,
despite the fact that this is the situation in the real world.
}

\section{Causality and Responsibility}\label{sec:resp}

\subsection{Causality}

We start with the definition of cause from \cite{HP01b}.  

\begin{definition}[Cause]\label{def-cause}
We say that $\vec{X} = \vec{x}$ is a {\em cause\/} of $\varphi$ in
$(M,\vec{u})$ if the following three conditions hold: 
\begin{description}
\item[AC1.] $(M,\vec{u}) \models (\vec{X} = \vec{x}) \wedge \varphi$. 
\item[AC2.] There exist a partition $(\vec{Z},\vec{W})$ of $\V$ with 
$\vec{X} \subseteq \vec{Z}$ and some setting 
$(\vec{x}',\vec{w}')$ of the
variables in $(\vec{X},\vec{W})$ such that if $(M,\vec{u}) \models Z = z^*$
for $Z \in \vec{Z}$, then
\be
\item[(a)] $(M,\vec{u}) \models [ \vec{X} \leftarrow \vec{x}',
\vec{W} \leftarrow \vec{w}']\neg{\varphi}$. That is, changing
$(\vec{X},\vec{W})$ from $(\vec{x},\vec{w})$ to 
$(\vec{x}',\vec{w}')$
changes 
$\varphi$ from \bft \ to \bff.
\item[(b)] $(M,\vec{u}) \models [ \vec{X} \leftarrow \vec{x},
\vec{W} \leftarrow \vec{w}', \vec{Z}' \leftarrow \vec{z}^*]\varphi$ for
all subsets $\vec{Z}'$ of $\vec{Z}$. That is, setting $\vec{W}$ to $\vec{w}'$
should have no effect on $\varphi$ as long as $\vec{X}$ has the value 
$\vec{x}$, even if all the variables in an arbitrary subset of $\vec{Z}$
are set to their original values in the context $\vec{u}$.  
\ee
\item[AC3.] $(\vec{X} = \vec{x})$ is minimal, that is, no subset of
$\vec{X}$ satisfies AC2.
\end{description}
\end{definition}

AC1 just says that $A$ cannot be a cause of $B$ unless both $A$ and $B$
are true, while AC3 is a minimality condition to prevent, for example,
Suzy throwing the rock and sneezing from being a cause of the bottle
shattering.   Eiter and Lukasiewicz \citeyear{EL01} showed that one
consequence of AC3 is that causes can always be taken to be single
conjuncts.  
Thus, from  here on in, we talk about $X=x$ being the cause of $\phi$,
rather than $\vec{X} = \vec{x}$ being the cause.
The core of this definition lies in AC2.
Informally, the variables in $\vec{Z}$ should be thought of as
describing the ``active causal process'' from $X$ to $\phi$.
These are the variables that mediate between $X$ and $\phi$.
AC2(a) is reminiscent of the traditional counterfactual criterion,
according to which $X=x$ is a cause of $\phi$ if change
the value of $X$ results in $\phi$ being false.
However, AC2(a) is more permissive than the traditional criterion;
it allows the dependence of $\phi$ on $X$ to be tested
under special {\em structural contingencies}, 
in which the variables $\vec{W}$ are held constant at some setting
$\vec{w}'$.  AC2(b) is an attempt to counteract the ``permissiveness''
of AC2(a) with 
regard to structural contingencies.  Essentially, it ensures that
$X$ alone suffices to bring about the change from $\phi$ to $\neg
\phi$; setting $\vec{W}$ to $\vec{w}'$ merely eliminates
spurious side effects that tend to mask the action of $X$.

To understand the role of AC2(b), consider the rock-throwing example
again.  In the model in Figure~\ref{fig0}, it is easy to see that both Suzy
and Billy are causes of the bottle shattering.  Taking $\vec{Z} = \{\ST,BS\}$,
consider the structural contingency where Billy doesn't throw ($\BT = 
0$).  Clearly $[\ST \gets 0, \BT \gets 0]BS = 0$ and $[\ST \gets 1, \BT 
\gets 0]BS = 1$ both hold, so Suzy is a cause of the bottle
shattering. A symmetric argument shows that Billy is also 
a
cause. 

But now consider the model described in Figure~\ref{fig1}.  It is still
the case that Suzy is a cause in this model.  We can take $\vec{Z} = \{\ST, \SH,
BS\}$ and again consider the contingency where Billy doesn't throw.
However, Billy is {\em not\/} a cause of the bottle shattering.  For
suppose that
we now take $\vec{Z} = \{\BT, \BH, BS\}$ and consider the contingency
where Suzy doesn't throw.  Clearly AC2(a) holds, since if Billy doesn't
throw (under this contingency), then the bottle doesn't shatter.
However, AC2(b) does not hold.  Since $\BH \in \vec{Z}$, if we set $\BH$ to
0 (it's original value), then AC2(b) requires that 
$[\BT \gets 1, \ST \gets 0, \BH \gets 0](BS = 1)$ hold, but it does not.
Similar arguments show that no other choice of $(\vec{Z},\vec{W})$ makes
Billy's throw a cause.

\stam{
Eiter and Lukasiewicz \cite{EL02} proved 
that for binary causal models, AC2 can be replaced by the following
condition (to get an equivalent definition of causality):
\begin{description}
\item[AC2$'$.] There exist a partition $(\vec{Z},\vec{W})$ of $\V$ with 
$\vec{X} \subseteq \vec{Z}$ and some setting $(\vec{x}',\vec{w}')$ of the
variables in $(\vec{X},\vec{W})$ such that if $(M,\vec{u}) \models Z = z^*$
for $Z \in \vec{Z}$, then
\be
\item $(M,\vec{u}) \models [ \vec{X} \leftarrow \vec{x}',
\vec{W} \leftarrow \vec{w}']\neg{\varphi}$. 
\item $(M,\vec{u}) \models [ \vec{X} \leftarrow \vec{x},
\vec{W} \leftarrow \vec{w}', \vec{Z} \leftarrow \vec{z^*}]\varphi$.
\ee
\end{description}
That is, for binary causal models it is enough to check that 
changing the value of $\vec{W}$ does not falsify $\varphi$ if all 
other variables keep their original values. 
}

In \cite{HP01b}, a slightly more refined definition of causality is also
considered, where there is a set of {\em allowable settings} for the
endogenous settings, and
the only contingencies that can be considered in AC2(b) are ones where
the settings $(\vec{W} = \vec{w}', \vec{X} = \vec{x}')$ and 
$(\vec{W} = \vec{w}', \vec{X} = \vec{x})$ are allowable.  The intuition
here is that we do not want to have causality demonstrated by an
``unreasonable'' contingency.  For example, in the rock throwing
example, we may not want to allow a setting where $\ST=0$, $\BT=1$, and
$\BH=0$, since this means that Suzy doesn't throw, Billy does, and yet
Billy doesn't hit the bottle; this setting contradicts the
assumption that Billy's throw is perfectly accurate.  We return to this
point later.

\stam{
AC2 allows arbitrary contingencies (i.e., setting of $\vec{W}$) to be
considered.  There are examples (e.g., \cite[Example 5.2]{HP01b},
\cite{HopkinsP03}) 
showing that this liberality may occasionally lead to inappropriate
results.  There are two restrictions we can consider on contingencies:%
\footnote{We remark that these restrictions are not discussed in the
current version of \cite{HP01b}; they will be discussed in a later
version.} 
\begin{itemize}
\item the only settings $\vec{x'}$ and $\vec{w'}$ that can be considered
in AC2(b) are ones such that, for some contexts $\vec{u}$ and
$\vec{u}'$, it is the 
case 
that $(M,\vec{u}) \sat \vec{X} = \vec{x} \land \vec{W} = \vec{w'}$ and
$(M,\vec{u'}) \sat \vec{X} = \vec{x'} \land \vec{W} = \vec{w'}$.  That
is, we do not consider settings in AC2(b) that do not arise in some
context.  
\item The only settings $\vec{w}'$ that we consider in AC2(b) are such
that, for any subset $\vec{W}''$ of $\vec{W}'$, if $\vec{w}''$ is the
subset of $\vec{w}'$ corresponding to $\vec{W}''$, we have
$(M,\vec{u}) \sat [\vec{W}'' \gets \vec{w}'']\phi$.  That is, we only
consider contingencies that have no effect on $\phi$.  (This restriction
was first proposed by Hitchcock \cite{Hit99}.)
\end{itemize}
We say that $\vec{X} = \vec{x}$ is a {\em justified cause\/} of
$\varphi$ in $(M,\vec{u})$ if AC1--3 hold and, in addition, the
contingencies in AC2 satisfy the two conditions above.
}

\subsection{Responsibility}

The definition of responsibility in causal models extends the definition
of causality. 
\begin{definition}[Degree of Responsibility]\label{def-resp}
The {\em degree of responsibility
of $X=x$ for $\phi$ in 
$(M,\vec{u})$\/}, denoted $\dr((M,\vec{u}), (X=x), \phi)$, is
$0$ if $X=x$ is 
not a cause of $\phi$ in $(M,\vec{u})$; it is $1/(k+1)$ if
$X=x$ is  a cause of $\phi$ in $(M,\vec{u})$ 
and there exists a partition $(\vec{Z},\vec{W})$ and setting
$(x',\vec{w}')$ for which AC2 holds
such that (a) $k$ variables in $\vec{W}$ have different values in $\vec{w}'$
than they do in the context $\vec{u}$ and (b) there is no partition 
$(\vec{Z}',\vec{W}')$ and setting $(x'',\vec{w}'')$ satisfying AC2
such that only $k' < k$ variables have different values in $\vec{w}''$
than they do the context $\vec{u}$.
\end{definition}

Intuitively, $\dr((M,\vec{u}), (X=x), \phi)$ measures the minimal number 
of changes that have to be made in $\vec{u}$ 
in order to make $\phi$ counterfactually depend on $X$.
If no partition of $\V$ to
$(\vec{Z},\vec{W})$ makes $\phi$ counterfactually depend on 
$(X=x)$, then the minimal number of changes in $\vec{u}$ 
in Definition~\ref{def-resp} is taken to have cardinality
$\infty$, and thus the degree of responsibility of $X=x$ is $0$. 
If $\phi$ counterfactually depends on $X=x$, that is, changing the
value of $X$ alone falsifies $\phi$ in $(M,\vec{u})$, then
the degree of responsibility of $X=x$ in $\phi$ is $1$. In other
cases the degree of responsibility is strictly between $0$ and $1$.
Note that $X=x$ is a {\em cause\/} of $\phi$ iff the degree of 
responsibility of $X=x$ for $\phi$ is greater than 0.

\begin{example} Consider the voting example from the
introduction.  Suppose there are 11 voters.  Voter $i$ is represented
by a variable $X_i$, $i = 1, \ldots, 11$; the outcome is represented by
the variable $O$, which is 
$1$ if Mr.~B wins and $0$ if Mr.~B wins.  
In the context where Mr.~B wins 11--0, it is easy to
check that each voter is
a cause of the victory (that is $X_i = 1$ is a cause of $O=1$, for $i =
1, \ldots, 11$).  However, the degree of responsibility of $X_i = 1$ for
is $O=1$ is just $1/6$, since at least five other voters must change
their votes before changing $X_i$ to 0 results in $O = 0$.  But now
consider the context where Mr.~B wins 6--5.  Again, each voter who votes
for Mr.~B is a cause of him winning.  However, now each of these voters
have degree of responsibility 1.  That is, if $X_i = 1$, changing $X_i$
to 0 is already enough to make $O = 0$; no other variables need to
change.
\end{example}

\begin{example}
It is easy to see that 
Suzy's throw has degree of responsibility $1/2$ for the bottle
shattering in the naive model described in Figure~\ref{fig0};
Suzy's throw becomes critical in the contingency where Billy does not
throw.  
In the ``better'' model of Figure~\ref{fig1}, Suzy's degree of
responsibility is 1.  If we take $\vec{W}$ to consist of $\{\BT, \BH\}$,
and keep both variables at their current setting (that is, consider the
contingency where $\BT=1$ and 
$\BH=0$), 
then Suzy's throw becomes
critical; if she throws, the bottle shatters, and if she does not throw,
the bottle does not shatter (since $\BH=0$).  As we suggested earlier,
the setting $(\ST=0,\BT=1,\BH=0)$ may be a somewhat unreasonable one to
consider, since it requires Billy's throw to miss.
If the setting  $(\BH=1, \ST=0, \BH=0)$ is not allowable, 
then we cannot consider this contingency.
In that case, Suzy's degree of responsibility is again $1/2$, since we
must consider the contingency where Billy does not throw.
Thus, the restriction to allowable settings allows us to capture what
seems like a significant intuition here.
\end{example}

As we mentioned in the introduction, 
in a companion paper \cite{CHK} we apply
our notion of responsibility to program verification.  The idea
is to determine the degree of responsibility of the setting 
of each state for the satisfaction of a specification 
in a given system.  
For example, given a specification
of the form $\Diamond p$ (eventually $p$ is true), if $p$ is true in
only one state of the verified system, 
then that state has degree of responsibility 1 for the
specification.  On the other hand, if $p$ is true in three states, each
state only has degree of responsibility $1/3$.  Experience has shown
that if there are many states with low degree of responsibility for a
specification, then either the specification is incomplete (perhaps $p$
really did have to happen three times, in which case the 
specification should
have said so), or there is a problem with the 
system generated by the program, since it has
redundant states.  

The degree of responsibility can also be used to provide a measure of
the degree of fault-tolerance in a system.  If a component is critical
to an outcome, it will have degree of responsibility 1.  To ensure 
fault tolerance, we need to make sure that no component has a high
degree of responsibility for an outcome.
Going back to the example of $\Diamond p$, the degree of responsibility
of $1/3$ for a state means that the system is robust to 
the simultaneous failures of at most two states.

\subsection{Blame}\label{sec:blame}

\stam{
\begin{definition}
Let $M$ be a model, $X$ a variable of $M$ and 
$x$ an assignment to $X$.
Let $\zug{U,W}$ be an epistemic state of $(X=x)$. The {\em blame\/}
of $(X=x)$ in $\varphi$ in the model $M$ relative to the epistemic 
state $\zug{U,W}$ is the sum of probabilities of contexts 
$\vec{u} \in U$ that fulfill the following conditions.
\be
\item $(M,\vec{u}) \models \neg{\varphi}$.
\item $(M,\vec{u}) \models [X=x]\varphi$.
\ee 
\end{definition}
}

The definitions of both causality and
responsibility  assume that the context and the structural equations are
given; there is no uncertainty.  We are often interested in assigning a
degree of {\em blame\/} to an action.  
This assignment depends on the
epistemic state of the agent {\em before\/} the action was performed.
Intuitively, if the agent had no reason to believe, 
before he performed the action,
that his action would
result in a particular outcome, 
then he should not be held to blame for the outcome
(even if in fact his action caused the outcome).  

\stam{
To deal with the fact that we are considering two points in
time---before the action was performed and after---we 
add superscripts to variables.  We
use a superscript 0
to denote the value of the random variable before the action was
performed and the superscript 1 to denote the value of the random
variable after.  Thus, $Y^0 = 1$ denotes that the random variable $Y$
has value 1 before the action is performed, while $Y^1 = 2$ denotes
that it had value 1 afterwards.  If $\phi$ is a Boolean combination of 
(unsuperscripted) 
random variables, we use $\phi^0$ and $\phi^1$ to denote the value of
$\phi$ before and after the action is performed, respectively.
}
There are two significant sources of uncertainty for an agent who is
contemplating performing an action:
\begin{itemize}
\item what the true situation is (that is, what value various variables
have); for example, a doctor may be 
uncertain about whether a patient has high blood pressure.  
\item how the world works; for example, a doctor may be uncertain about
the side effects of a given medication;
\end{itemize}

In our framework, the ``true situation'' is determined by the context;
``how the world works'' is determined by the structural equations.
Thus, we model an agent's uncertainty by a pair $(\K,\Pr)$, where $\K$
is a set of pairs of the form $(M,\vec{u})$,
where $M$ is a 
causal model and $\vec{u}$ is a context,
and $\Pr$ is a probability distribution over $\K$.
Following \cite{HP01a}, who used such epistemic states in the
definition of {\em explanation}, we call a pair $(M,\vec{u})$ a {\em 
situation}.  
We think of $\K$ as describing the situations that the agent considers
possible before $X$ is set to $x$.
The degree 
of blame that setting $X$ to $x$ has for $\phi$ is 
then
the expected degree
of responsibility of $X=x$ for $\phi$ in 
$(M_{X \gets x},\vec{u})$,   
taken over the situations $(M,\vec{u})
\in \K$. 
Note that the situation $(M_{X \gets x},\vec{u})$ for $(M,
\vec{u}) \in \K$ are those 
that the agent considers possible after $X$ is set to
$x$. 
\begin{definition}[Blame]\label{def:blame}
The {\em degree of 
blame of setting $X$ to $x$ for $\phi$ relative to epistemic state
$(\K,\Pr)$\/}, denoted $\db(\K,\Pr,X \gets x, \phi)$, is
$$\sum_{(M,\vec{u}) \in \K}
\dr((M_{X \gets x}, \vec{u}), X = x, \phi)
\Pr((M,\vec{u})).$$ 
\end{definition}

\begin{example} Suppose that we are trying to compute the degree
of blame of Suzy's throwing the rock for the bottle shattering. 
Suppose that the only causal model that Suzy considers possible is
essentially like that of Figure~\ref{fig1}, with some minor modifications:
$\BT$ can now take on three values, say 0, 1, 2; as before, if
$\BT = 0$ then Billy doesn't throw, if $\BT = 1$, then Billy does throw,
and if $\BT = 2$, then Billy throws extra hard.  Assume that the causal
model is such that if $\BT = 1$, then Suzy's rock will hit the bottle
first, but if $\BT =2$, they will hit simultaneously.  Thus, $\SH = 1$
if $\ST = 1$, and $\BH = 1$ if $\BT = 1$ and $\SH = 0$ or if $\BT = 2$.
Call this structural model $M$.

At time 0, Suzy considers the following four situations equally likely:
\begin{itemize}
\item $(M, \vec{u}_1)$, where $\vec{u}_1$ is such that Billy already
threw at time 0 (and hence the bottle is shattered);
\item $(M,\vec{u}_2)$, where the bottle was whole before
Suzy's throw, 
and Billy throws extra hard, so 
Billy's throw and Suzy's throw hit the bottle
simultaneously (this essentially gives the model in Figure~\ref{fig0});
\item $(M,\vec{u}_3)$, where the bottle was whole before Suzy's throw,
and Suzy's throw hit before Billy's throw (this essentially gives the
model in Figure~\ref{fig1}); and
\item $(M,\vec{u}_4)$, where the bottle was whole before
Suzy's throw, and Billy did not throw. 
\end{itemize}
The
bottle is already shattered in $(M,\vec{u}_1)$ before Suzy's
action,
so Suzy's throw is not a cause of the bottle shattering, and her degree
of responsibility for the shattered bottle is 0.
As discussed earlier,
the degree of 
responsibility of Suzy's throw for the bottle shattering is $1/2$ in
$(M,\vec{u}_2)$ and 1 in both $(M,\vec{u}_3)$ and ($(M,\vec{u}_4)$.
Thus, the degree of blame is $\frac{1}{4}\cdot \frac{1}{2} +
\frac{1}{4}\cdot 1 + \frac{1}{4}\cdot 1 = \frac{5}{8}$.  
If we further require that the contingencies in AC2(b) involve only
allowable settings, and assume that the setting $(\ST=0, \BT=1, \BH=0)$ is not
allowable, then the degree of responsibility of Suzy's throw in
$(M,\vec{u}_3)$ is $1/2$; in this case, the degree of blame is
$\frac{1}{4}\cdot \frac{1}{2} + \frac{1}{4}\cdot \frac{1}{2} +
\frac{1}{4}\cdot 1 = \frac{1}{2}$.  
\end{example}

\stam{
To understand why,
suppose that we are trying to compute the degree of blame of Suzy's
turning off the fridge for the ice-cream melting.  Suppose that Suzy
considers the following three situations possible before she turns off the
fridge: 
\begin{itemize}
\item $s_1$, where the fridge is already turned off;
\item $s_2$, where the ice cream is already melted;
\item $s_3$, where the fridge is on and the ice cream is hard.
\end{itemize}
Further suppose that Suzy considers these three situations to be equally
likely.  To compute the degree of blame assigned to Suzy's turning off the
fridge for the ice cream melting, we ignore $s_1$ and $s_2$ (because the
fridge is already turned off in $s_1$ and the ice cream is already
melted in $s_2$).  If the fridge being off is the only cause of the ice
cream melting in the situation $s_3$, then the degree of blame for the
ice cream melting is $1/3$.  Of course, in $s_3$, the degree of
responsibility of turning off the fridge for the ice cream melting is 1.
}

\begin{example} 
Consider again the example of the firing squad with ten 
excellent marksmen.  Suppose that marksman 1 knows that exactly one
marksman has a live bullet in his rifle, 
and that all the marksmen will shoot.
 Thus, he considers 10
augmented
situations possible, depending on who has the bullet.  Let $p_i$ be his
prior probability that marksman $i$ has the live bullet.  Then the
degree of blame of his shot for the death is $p_i$.  The degree of
responsibility is either 1 or 0, depending on whether or not he actually
had the live bullet.  Thus, it is possible for the degree of
responsibility to be 1 and the degree of blame to be 0 (if he 
mistakenly
ascribes probability 0 to his having the live bullet, when in fact he does), and
it is possible for the degree of responsibility to be 0 and the degree
of blame to be 1 (if he mistakenly ascribes probability 1 to 
his having the bullet when he in fact does not).
\end{example}

\begin{example}
The previous example suggests that both degree of blame and degree of
responsibility may be relevant in a legal setting.  Another issue that
is relevant in legal settings is whether to consider actual epistemic
state or to consider what the epistemic state should have been.  
The former is relevant when considering intent.  To see the relevance of
the latter, 
consider a patient who dies as a result of being treated by a doctor
with a particular drug.
Assume that the patient died due to the drug's adverse side effects 
on people with high blood pressure and, for simplicity, that this was
the only cause of death.Suppose that the doctor was not 
aware of the drug's adverse side
effects.  (Formally, this means that he does not consider possible a
situation with a causal model where taking the drug causes death.)
Then, relative to the doctor's actual epistemic state, 
the doctor's degree of blame will be 0.
However, a lawyer might argue in court that the doctor should  have
known that treatment had adverse side effects for patients with high
blood pressure (because this is well documented in the literature) 
and thus should have checked the patient's blood pressure.
If the doctor had performed this test, he would of course have known
that the patient had high blood pressure.  With respect to the resulting
epistemic state, the doctor's degree of blame for the death is quite
high.  Of course, the lawyer's job is to convince the court that the
latter epistemic state is the appropriate one to consider when assigning
degree of blame.  
\end{example}

Our definition of blame considers the epistemic state of the agent {\em
before\/} the action was performed.  It is also of interest to consider
the expected degree of responsibility {\em after\/} the action was
performed.  To understand the differences, again consider consider the
patient who dies as a result of being treated by a doctor with a
particular drug.  The doctor's epistemic state after the patient's death
is likely to be quite different from her epistemic state before the
patient's death.  She may still consider it possible that the patient
died for reasons other than the treatment, but will consider causal
structures where the treatment was a cause of death more likely.  Thus,
the doctor will likely have higher degree of blame relative to her 
epistemic state after the treatment. 

Interestingly, all three epistemic states (the epistemic state that an
agent actually has before performing an action, the epistemic state that
the agent should have had before performing the action, and the
epistemic state after performing the action) have been considered
relevant to determining responsibility according to different legal theories
\cite[p.~482]{HH85}.

\section{The Complexity of Computing Responsibility and
Blame}\label{sec:complexity} 

In this section we present complexity results for 
computing the degree
of responsibility and blame for general recursive models. 

\subsection{The complexity of computing responsibility}

Complexity results for computing causality were presented by Eiter and
Lukasiewicz \citeyear{EL02,EL01}.  They showed that 
the problem of detecting whether $X=x$ is an actual cause 
of $\varphi$ is 
$\Sigma_2^P$-complete for general recursive models and 
NP-complete for binary models \cite{EL01}. 
(Recall that $\Sigma_2^P$ is the second level of the polynomial hierarchy
and that binary models are ones where all random variables can take on
exactly two values.) 
There is a similar gap between the complexity of computing the degree of
responsibility and blame in general models and in binary models.

For a complexity class $A$, \fpa\ consists of all 
functions that can be computed 
by a polynomial-time Turing machine with an oracle for a problem in
$A$, which on input $x$ asks a total of $O(\log{|x|})$ queries 
(cf.~\cite{Pap84}). 
We show that computing the degree of responsibility 
of $X=x$ for $\phi$ in arbitrary models is \fpsigma-complete.
It is shown in \cite{CHK} 
that computing the degree of
responsibility in binary models is \fp-complete.
\stam{
The computation of blame can be performed by computing the degree of responsibility
separately for each $(M,\vec{u})$ in $\K$ and then computing the expected degree
of responsibility according to the probability function $\Pr$. 
This involves running the algorithm for responsibility $|\K|$ times, and thus posing
$|\K|\log{n}$ queries to the $\Sigma^P_2$ oracle (NP oracle for binary models). 
} %

Since there are no known natural \fpsigma-complete problems, 
the first step in showing that computing the degree of responsibility is
\fpsigma-complete is to define an \fpsigma-complete problem.
We start by defining one that we call MAXQSAT$_2$. 

Recall that 
a {\em quantified Boolean formula} \cite{Stock} (QBF) has 
the form $\forall X_1 \exists X_2 \ldots \psi$, where
$X_1, X_2, \ldots$ are propositional variables and $\psi$ is a
propositional formula.  A QBF is {\em closed\/} if it has no free
propositional variables. 
TQBF consists of the closed QBF formulas that are
true.  For example, $\forall X \exists Y (X \Rightarrow Y) \in {\rm TQBF}$.
As shown by Stockmeyer \citeyear{Stock}, the following problem 
QSAT$_2$ is $\Sigma_2^P$-complete:
\[
\begin{array}{c}
{\rm QSAT}_2 = \{ \exists X \forall Y \psi(X,Y) 
\in {\rm TQBF}:
\ \psi \in 3{\rm -CNF} 
 \}. 
\end{array}
\]
That is, QSAT$_2$ is the language of all true 
QBFs of the form 
$\exists \vec{X} \forall \vec{Y} \psi$, 
where $\psi$ is a Boolean formula in 3-CNF.

A {\em witness\/} $f$ for a true closed QBF 
$\exists \vec{X} \forall \vec{Y} \psi$ is an assignment $f$ to $\vec{X}$
under which $\forall \vec{Y} \psi$ is true.  
We define MAXQSAT$_2$ as the problem of computing the maximal number of 
variables in $\vec{X}$ that can be assigned $1$ in 
a witness for $\exists \vec{X} \forall \vec{Y} \psi$.
Formally, given a QBF 
$\Phi = \exists \vec{X} \forall \vec{Y} \psi$, 
define MAXQSAT$_2(\Phi)$ to be $k$ if
there exists 
a witness for $\Phi$ that assigns exactly $k$ of the variables in
$\vec{X}$ the value $1$, and every other witness for $\Phi$ assigns
at most $k' \leq k$ variables in $\vec{X}$ the value $1$. If
$\Phi \notin$ QSAT$_2$, then MAXQSAT$_2(\Phi) = -1$.

\begin{theorem}\label{theor-maxqsat}
MAXQSAT$_2$ is \fpsigma-complete.
\end{theorem}
\begin{proof}
First we prove that MAXQSAT$_2$ is 
in \fpsigmas by describing an algorithm in \fpsigmas for solving 
MAXQSAT$_2$. The algorithm queries an oracle
$O_\Lan$ for the language $\Lan$, defined as follows:
\[
\begin{array}{c}
\Lan = \{ (\Phi,k) : k \ge 0,\,
{MAXQSAT_2}(\Phi) \geq k \}.
\end{array}
\] 
It is easy to see that $\Lan \in \Sigma_2^P$:
if $\Phi$ has the form $\exists \vec{X} \forall \vec{Y} \psi$, guess
an assignment $f$ that assigns at least $k$ variables in 
$\vec{X}$ the value $1$ and check 
whether $f$ is a witness for $\Phi$.
Note that this amounts to checking the validity of
$\psi$ with each variable in $\vec{X}$ replaced by its value according
to $f$, so this check is in co-NP, as required.
It follows that 
the language $\Lan$ is in $\Sigma_2^P$. 
(In fact, it is only a slight variant of QSAT$_2$.) 
Given $\Phi$, the
algorithm for computing MAXQSAT$_2(\Phi)$ 
first checks if $\Phi \in \mbox{QSAT}_2$ by making 
a query with $k=0$.
If it is not, then MAXQSAT$_2(\Phi)$ is $-1$.  If
it is, the algorithm
performs a binary search for its value,
each time dividing the range of possible values by $2$ 
according to the answer of $O_\Lan$. The number of possible 
values of MAXQSAT$_2(\Phi)$ is then 
$|X| + 2$ (all values between $-1$ and $|X|$ are possible),
so the algorithm asks $O_\Lan$ at most 
$\lceil \log{n} \rceil + 1$ 
queries.

Now we show that MAXQSAT$_2$ is \fpsigma-hard
by describing a generic reduction from
a problem in \fpsigmas to MAXQSAT$_2$. 
Let $f:\{0,1\}^* \rightarrow \{0,1\}^*$ be
a function in \fpsigma. That is, there exists a deterministic oracle
Turing machine $M_f$ that on input $w$ outputs $f(w)$ for each 
$w \in \Sigma^*$, operates in time at most $|w|^c$ for a constant $c
\geq 0$, and 
queries an oracle for the language QSAT$_2$ at most 
$d\log{|w|}$ times (for all sufficiently large inputs $w$)
during the execution. We now describe a reduction from $f$ to MAXQSAT$_2$. 
Since this is a reduction between function problems, we have to give 
two polynomial-time functions $r$ and $s$ such that for every input
$w$, we have that $r(w)$ is a QBF and $s({\rm MAXQSAT}_2(r(w))) = f(w)$.  

We start by describing a deterministic polynomial Turing machine $M_r$
that on input $w$ computes $r(w)$. On input $w$ of length $n$, $M_r$ starts
by simulating $M_f$ on $w$. At some step during the simulation $M_f$
asks its first oracle query $q_1$. The query $q_1$ is a QBF 
$\exists \vec{Y}_1 \forall \vec{Z}_1 \psi_1$. The machine $M_r$ cannot figure
out the answer to $q_1$, thus it writes $q_1$ down and continues with the 
simulation. Since $M_r$ does not know the answer, it has to simulate the run
of $M_f$ for both possible answers of the oracle.
The machine $M_r$ continues in this fashion, keeping track of all 
possible executions of $M_f$ on $w$ for each sequence of answers to the oracle
queries. Note that a sequence of answers to oracle
queries uniquely characterizes a specific execution of $M_f$ (of length 
$n^c$). 
Since there are $2^{d\log{n}} = n^d$ possible sequences of answers,
$M_r$ on $w$ has to simulate $n$ possible executions of $M_f$ on $w$,
thus the running time of this step is bounded by $O(n^{c+d})$. 

The set of possible executions of $M_f$ can be viewed as a tree of
height $d \log {n}$ (ignoring the computation between queries to the
oracle).  There are at most $n^{d+1}$ queries on this tree.  These
queries have the form 
$\exists \vec{Y}_i \forall \vec{Z}_i \psi_i$, 
for
$i = 1, \ldots, n^{d+1}$. 
We can assume without loss of generality that these formulas involve pairwise
disjoint sets of variables. 

Each execution 
of $M_f$ on $w$ involves at most $d\log{n}$ queries from this collection.
Let $X_j$ be a variable 
that
describes the answer to the $j$th
query in an  execution, for $1 \leq j \leq d\log{n}$. 
(Of course, which query $X_j$ is the answer to depends on the
execution.)
Each of the $n^d$ possible assignments to the variables $X_1, \ldots,
X_{d\log{n}}$ can be thought of as representing a number $a \in \{0, \ldots,
n^d-1\}$ in binary. 
Let $\zeta_a$ be the formula that characterizes the assignment to these
variables corresponding to the number $a$.  That is, 
$\zeta_a = \wedge_{j=1}^{d\log{n}} X^a_j$, where $X^a_j$ is 
$X_j$ if
the $j$th bit of $a$ is $1$ in $a$ and 
$\neg{X_j}$ otherwise. 
Under the interpretation of $X_j$ as determining the answer the query
$j$, each such assignment $a$ determines an execution of $M_f$.  
Note that the assignment corresponding to the highest number $a$ for which
all the queries corresponding to bits of $a$ that are 1 are true is
the one that corresponds to the actual execution of $M_f$.
For suppose that this is not the case. That is, suppose that the actual
execution 
of $M_f$ corresponds to some $a' < a$. Since $a' < a$, there exists
a bit $X_j$ that is $1$ in $a$ and is $0$ in $a'$. By our choice of 
$a$, the query $\exists \vec{Y}_i \forall \vec{Z}_i \psi$ corresponding
to $X_j$ is a true QBF, thus a $\Sigma^P_2$-oracle that $M_f$ queries should
have answered ``yes'' to this query, and we reach a contradiction.

The next formula provides 
the
connection between the $X_i$s and the queries.
It says that if $\zeta_a$ is true, then all the queries corresponding to
the bits that are 1 in $a$ must be true QBF.
For each assignment $a$, let $a_1, \ldots,
a_{N_a}$ be the queries that were answered ``YES'' during the execution
of the $M_f$ corresponding to $a$.  Note that $N_a \le d \log{n}$.
Define
\[ 
\begin{array}{c}
\eta = \bigwedge_{a=0}^{n^d-1} (\zeta_a \Rightarrow
\exists \vec{Y}_{a_1} \forall \vec{Z}_{a_1} \psi_{a_1} \wedge \ldots
\exists \vec{Y}_{a_{N_a}} \forall \vec{Z}_{a_{\log{n}}} \psi_{a_{N_a}}
). 
\end{array} \]

The variables $\vec{Y}_{a_i}$, $\vec{Z}_{a_i}$ do not appear in the formula
$\eta$ outside of $\psi_{a_i}$, thus we can re-write $\eta$ as
\[ \eta = \exists \vec{Y}_1, \ldots, \vec{Y}_{n^d}, \ldots,
          \forall \vec{Z}_1, \ldots, \vec{Z}_{n^d}  \eta', \]
where
\[ \eta' = \bigwedge_{a=0}^{n^d-1} \zeta_a \Rightarrow 
           (\psi_{a_1} \wedge \ldots \wedge \psi_{a_{\log{n}}}). \]

The idea now is to use MAXQSAT$_2$ to compute the highest-numbered
assignment $a$ such that all the queries corresponding to bits that are
1 in $a$ are true.  To do this,  it is useful to express the number $a$
in unary.  Let $U_1, \ldots, U_{n^d}$ be fresh variables.  Note that
$U_1, \ldots, U_{n^d}$ represent a number in unary if
there exists $1 \leq j \leq n$ such that $U_i = 1$ for each $i < j$ and
$U_i = 0$ for each $i \geq j$. 
Let $\xi$ express the fact that the $U_i$s 
represent
the same number as $X_1, \ldots, X_{d\log n}$, but in unary:
$$\xi=\bigwedge_{i=1}^{n^d} (\neg{U_i} \Rightarrow \neg{U_{i+1}}) \bigwedge
(\neg{U_1} \Rightarrow (X^0_1 \wedge \ldots \wedge X^0_{\log n})) 
\wedge \bigwedge_{a=1}^{n^d-1} [ (U_a \wedge \neg{U_{a+1}}) \Rightarrow 
(X^a_1 \wedge \ldots \wedge X^a_{d\log n})].$$

Let $\Phi = \exists U_1, \ldots, U_{n^d}, X_1, \ldots, X_{d\log{n}}
 (\eta \wedge \xi)$.
Note that the size of $\Phi$ is polynomial in $n$ and, 
for each $i$, the variables $\vec{Y}_i$ and $\vec{Z}_i$ 
do not appear in $\varphi$ outside of $\eta$. Thus, we can
rewrite $\Phi$ in the following form:
\[
\begin{array}{c}
\Phi = \exists U_1, \ldots, U_{n^d}, X_1, \ldots, X_{d\log{n}}, 
       \vec{Y}_1, \ldots, \vec{Y}_{n^d} 
       \forall \vec{Z}_1, \ldots, \vec{Z}_{n^d} (\eta' \wedge \xi).
\end{array}
\]
Thus, $\Phi$ has the form $\exists \vec{X} \forall \vec{Y} \psi$.

We are interested in the 
witness for $\Phi$ that makes the most $U_i$'s true, since this will
tell us the execution actually followed by $M_f$.  
While MAXQSAT$_2$ gives us 
the maximal number of variables that can be assigned true in
a witness for $\Phi$, 
this is not quite the information we need. 
Assume for now that we have a machine that is capable of
computing a slightly generalized version of MAXQSAT$_2$. We denote this
version by SUBSET\_MAXQSAT$_2$ and define it as a maximal number of 
$1$'s in a witness computed for a given subset of variables (and not
for the whole set as MAXQSAT$_2$).  Formally, given a QBF 
$\Psi=\exists \vec{X} \forall \vec{Y} \psi$ and a set 
$\vec{Z} \subseteq \vec{X}$, we define SUBSET\_MAXQSAT$_2(\Psi,\vec{Z})$
as the maximal number of variables from $\vec{Z}$ assigned $1$ in a
witness for $\Psi$, or $-1$ if $\Psi$ is not in TQBF. 

Clearly, SUBSET\_MAXQSAT$_2(\Phi,\vec{U})$ gives us the assignment which
determines the execution of $M_f$.  Let 
$r(w) = \Phi$.  Let $s$ be the function that extracts
$f(w)$ from SUBSET\_MAXQSAT$_2(\Phi, \{ U_1, \ldots, U_n \})$.
(This can be done in polynomial time, since $f(w)$ can be computed in
polynomial time, given the answers to the oracle.)

\stam{
We claim that SUBSET\_MAXQSAT$_2(\Phi, \{ U_1, \ldots, U_n \} )$
corresponds to the correct value of 
$f(w)$. Let $f_{max}$ be the 
witness for $\Phi$ that assigns 
SUBSET\_MAXQSAT$_2(\Phi, \{ U_1, \ldots, U_n \} )$ variables out of
$U_1, \ldots, U_n$ the value $1$,
and let $a$ be the sequence of answers to oracle queries derived from
the assignment $f_{max}$.
First note that all positive answers for oracle queries 
derived from 
$f_{max}$ are correct. Indeed,
by the construction of $\Phi$, if $X_i$ is assigned $1$, for each
$0 \leq i \leq \log{n}$, then the corresponding query 
$\exists Y_{a_i} \forall Z_{a_i} \psi_{a_i}$
is a true sentence. Thus, the only possible error could be if 
we mistakenly assigned one of 
$X_i$s
the value $0$. Assume that this is
the case. Then, if we change the value of 
$X_i$
from $0$ to $1$
we still get a satisfying assignment $f'$. However, in $f'$ the number of
variables from $U_1, \ldots, U_n$ assigned $1$ is greater than in 
$f_{max}$, in the contrary to the definition of $f_{max}$. 

Thus, $r(w) = \Phi$, and $s$ extracts the value of
$f(w)$ from SUBSET\_MAXQSAT$_2(\Phi, \{ U_1, \ldots, U_n \})$.
}
It remains to show how to reduce SUBSET\_MAXQSAT$_2$ to MAXQSAT$_2$.
Given a formula $\Psi = \exists \vec{X} \forall \vec{Y} \psi$ and
a set $\vec{Z} \subseteq \vec{X}$, we compute SUBSET\_MAXQSAT$_2(\Psi,\vec{Z})$
in the following way. Let $\vec{U} = \vec{X} \setminus \vec{Z}$. For each
$U_i \in \vec{U}$ we add a variable $U'_i$. Let $\vec{U}'$ be the set of all
the $U_i'$ variables.
Define the formula $\Psi'$ as
$\exists \vec{X} \vec{U}' \forall \vec{Y} (\psi \wedge 
\bigwedge_{U_i \in U} (U_i \Leftrightarrow \neg{U'_i}))$. Clearly, in
all consistent 
assignments to $\vec{X} \cup \vec{U}'$ exactly half of the variables
in $\vec{U} \cup \vec{U}'$ are assigned $1$. Thus, the witness that
assigns MAXQSAT$_2(\Psi')$ variables the value $1$ also assigns 
SUBSET\_MAXQSAT$_2(\Psi,\vec{Z})$ variables from $\vec{Z}$ the value $1$.
The value SUBSET\_MAXQSAT$_2(\Psi,\vec{Z})$ is computed by subtracting
$|\vec{U}|$ from MAXQSAT$_2(\Psi')$. 
\end{proof}

Much like MAXQSAT$_2$, we define 
MINQSAT$_2(\exists \vec{X} \forall \vec{Y} \psi)$ to be the
minimum number of variables in $\vec{X}$ that can be assigned 1 in 
a witness for $\exists \vec{X} \forall \vec{Y} \psi$ if there is such
a witness, and $|\vec{X}| + 1$ otherwise.
This problem has the same complexity as MAXQSAT$_2$, as we 
now show.
\begin{lemma}\label{lemma-minqsat}
MINQSAT$_2$ is \fpsigma-complete. 
\end{lemma}
\begin{proof}
For a Boolean formula $\psi$ in 3-CNF, let $\overline{\psi}$ be
the same formula, except that each propositional variable $X$ is
replaced by its negation. It is easy to see
that $${\rm MAXQSAT}_2(\exists \vec{X} \forall\vec{Y}  \psi) = 
|\vec{X}| - {\rm MINQSAT}_2(\exists \vec{X} \forall \vec{Y} \overline{\psi}).$$
Thus, MINQSAT$_2$ and MAXQSAT$_2$ are polynomially reducible to
each other and therefore, by Theorem~\ref{theor-maxqsat},
MINQSAT$_2$ is \fpsigma-complete.
\end{proof}

We are now ready to prove \fpsigma-completeness of computing the degree
of responsibility for general recursive models. 

\begin{theorem}\label{theor-resp-general}
Computing 
the degree of responsibility is \fpsigma-complete in general recursive
models.  
 \end{theorem}
\begin{proof}
First we prove membership in \fpsigmas by describing an algorithm 
in \fpsigmas for computing the degree of responsibility. 
Consider the oracle
$O_\Lan$ for the language $\Lan$, defined as follows:
\[
\begin{array}{c}
\Lan = \{ \zug{(M,\vec{u}), (X=x), \varphi,i} : 
0 \geq i \geq 1, \; {\rm and \;} 
\dr((M,\vec{u}), (X=x), \varphi) \geq i \}. 
\end{array}
\]
In other words, $\zug{(M,\vec{u}), (X=x), \varphi,i} \in \Lan$ 
for $i > 0$ if 
there is a partition $(\vec{Z},\vec{W})$ and setting
$(x',\vec{w}')$ satisfying condition AC2 in Definition~\ref{def-cause}
such that  at most 
$1/i-1$ variables in $W$ have values different in $\vec{w}'$ from their
value in the context $\vec{u}$.
It is easy to see that $\Lan \in \Sigma_2^P$. 
Given as input a situation $(M,\vec{u})$, $X=x$ (where $X$ is an
exogenous variable in $M$ and $x \in \R(X)$), and a formula $\phi$,
the algorithm for computing the degree of responsibility uses the oracle
to perform a  
binary search on the 
value of $\dr((M,\vec{u}), (X=x), \varphi)$, each time
dividing the range of possible values for the degree of responsibility 
by $2$ according to the answer of $O_\Lan$. 
The number of possible candidates for the degree of responsibility is
$n-|X|+1$, where $n$ is the number of endogenous variables in $M$.
Thus, the algorithm runs in linear time (and logarithmic space) in the
size of the input and uses at most $\lceil \log{n} \rceil$ oracle
queries.  Note that the number of oracle queries depends only on the
number of endogenous variables in the model, and not on any other
features of the input.

The proof that computing the degree of responsibility is \fpsigma-hard 
essentially follows from an argument in \cite{EL02} 
showing that QSAT$_2$ can be reduced to the problem of detecting causality.
In fact, their argument actually provides  a reduction 
from MINQSAT$_2$ to the degree of responsibility. 
We now describe the reduction in more detail. Given as input a closed QBF
$\Phi = \exists \vec{A} \forall \vec{B} \varphi$, where $\varphi$ is
a propositional formula, define the situation 
$(M,\vec{u})$ as follows. 
$M = ((\U,\V,\R),\cF)$, where
\begin{itemize}
\item $\U$ consists of the single variable $E$; 
\item  $\V$ consists of the variables $\vec{A} \cup \vec{B} \cup \{C,X\}$, 
where $C$ and $X$ are fresh variables; 
\item $\R$ is defined so that 
each variable $S \in \vec{A} \cup \{ C,X \}$ has range $\R(S) = \{ 0,1 \}$,
while each variable $S \in \vec{B}$ has range $\R(S) = \{ 0,1,2 \}$;
\item $\cF$ is defined so that 
$F_S = C + X$ for $S \in \vec{B}$
and $F_S = 0$ for $S \in \V - \vec{B}$.
\end{itemize}
Let $\vec{u}$ be the context in which $E$ is set to 0.  
Note that every variable in $\V$ has value 0 in context $\vec{u}$. (This
would also be true in the context where $E$ is set to 1; the context
plays no role here.)
Let
\[ \psi = (\varphi' \wedge \bigwedge_{S \in \vec{B}} S \not = 2) \vee (C=0) 
\vee (X=1 \wedge C=1 \wedge \bigvee_{S \in \vec{B}} S \not = 2), \] 
where $\varphi'$ is obtained from $\varphi$ by replacing each variable
$S \in \vec{A} \cup \vec{B}$ in $\varphi$ by $S=1$.
It is shown in \cite{EL02} that $X=0$ is a cause of $\psi$ in
$(M,\vec{u})$ iff $\Phi \in {\rm TQBF}$. 
We now extend their argument to show that the degree of responsibility
of $X=0$ for $\psi$ is actually $1/({\rm MINQSAT}_2(\Phi)+2)$.

Assume first that $(X=0)$ is a cause
of $\psi$ in $(M,\vec{u})$. Then there exists a partition $(\vec{Z},\vec{W})$
of $\V$ and a setting $\vec{w}$ of $\vec{W}$ 
such that 
$(M,u) \models [X \gets 1, \vec{W} \gets \vec{w}]\neg\psi$ and
$(M,u) \models [X \gets 0, \vec{W} \gets \vec{w}, \vec{Z'} \gets \vec{0}]\psi$,
where $\vec{Z}'$ is as in AC2(b).  (Since every variable gets value 0 in 
context $\vec{u}$, the $\vec{z}^*$ of AC2(b) becomes $\vec{0}$ here.)
Suppose that $\vec{W}$ is a minimal set that satisfies these conditions.

Eiter and Lukasiewicz show that 
$\vec{B} \cap \vec{W} = \emptyset$.  To prove this, suppose that
$S \in \vec{B} \cap \vec{W}$.
If $S$ is set to either 0 or 1 in $\vec{w}$, then $(M,u) \not\models [\vec{W}
\gets \vec{w}, X \gets 1]\neg\psi$, 
since the clause $(X=1 \wedge C= 1 \wedge \bigvee_{S \in \vec{B}} S \not = 2)$
will be satisfied. 
On the other hand, 
if $S$ is set to 2 in $w$, then 
$(M,u) \not\models [\vec{W} \gets \vec{w}, X \gets 0]\psi$.
Thus, $\vec{B} \cap \vec{W} = \emptyset$.
It follows that $\vec{W} \subseteq \vec{A} \cup \{ C \}$, and so 
$\vec{Z} \subseteq {B} \cup \{ X \}$. 
Moreover, $C \in \vec{W}$, since
$C$ has to be set to $1$ in order to falsify $\psi$. 

Note that every variable in $\vec{W}$ is set to 1 in $\vec{w}$.  
This is obvious for $C$.  If there is some variable $S$ in $\vec{W}\cap
\vec{A}$ that is set to 0 
in $\vec{w}$, we could take $S$ out of $\vec{W}$ while still satisfying
condition AC2 in Definition~\ref{def-cause}, since every variable $S$ in
$\vec{W} \cap \vec{A}$ has value 0 in context $\vec{u}$ and can only be
1 if it is set to 1 in $\vec{w}$.  This contradicts the minimality of $W$.
To show that $\Phi \in {\rm TQBF}$, consider the truth assignment $v$ that
sets a variable $S \in \vec{A}$ true iff $S$ is set to 1 in $\vec{W}$.
It suffices to show that $v \models \forall \vec{B} \phi$.  Let $v'$ be
any truth assignment that agrees with $v$ on the variables in $\vec{A}$.
We must show that $v' \models \phi$.  
Let $\vec{Z}'$ consist of all variables
$S \in \vec{B}$ such that $v'(S)$ is false.  Since $(M,u) \models [X
\gets 0, \vec{W} \gets \vec{1}, \vec{Z'} \gets \vec{0}]\psi$, it follows
that $(M,u) \models [X \gets 0, \vec{W} \gets \vec{1}, \vec{Z}' \gets
\vec{0}]\phi'$.  It is easy to check that if every variable in $\vec{W}$ 
is set to 1, $X$ is set to 0, and $\vec{Z}'$ is set to
$\vec{0}$, then every variable in $\vec{A} \cup \vec{B}$ gets the same
value as it does in $v'$.  Thus, $v' \models \phi$.  It follows that
$\Phi \in {\rm TQBF}$.  This argument also shows that there is a
a witness
for $\Phi$ (i.e., valuation which makes $\forall \vec{B} \phi$ true)
that assigns 
the value true to all the variables in $\vec{W} \cap \vec{A}$, and only to these variables.

\stam{
Since $(M,\vec{u}) \models [X \gets 1,  \vec{W} \leftarrow \vec{w} ]\psi$,
we have that $(M,\vec{u}) \models [ \vec{W} \leftarrow \vec{w} ]\varphi'$.
Moreover, for all subsets $\vec{Z}'$ of $\vec{Z}$, 
$(M,\vec{u}) \models 
[ \vec{W} \leftarrow \vec{w}, \vec{Z} \leftarrow \vec{z}* ]\varphi'$,
where $\vec{z}*$ is value of $\vec{Z}'$ in $\vec{u}$. 
In other words, $\varphi$ is satisfied under the assignment $\vec{w}$ to
$\vec{W}$ for all possible assignments to $\vec{Z}$. Since 
$\vec{W} \subseteq \vec{A} \cup \{ C \}$, we have that $\Phi \in {\rm
TQBF}$ and 
$\vec{w}$ is the witness for the truth value of $\Phi$. Assume now that
$\vec{W}$ is the smallest set that satisfies the conditions of
Definition~\ref{def-cause}. Since $\vec{w}$ differs from $\vec{u}$,
we have that $\vec{w}$ assigns all variables in $\vec{W}$ the value $1$.
Then, by Definition~\ref{def-resp}, the degree of responsibility
$dr((M,\vec{u}),(X=0),\psi)$ of $X=0$ in the value of $\psi$ in
$(M,\vec{u})$ is $1/(|\vec{W}| + 1)$. On the other hand, $\vec{w}$
restricted to $\vec{A}$ is a witness for $\Phi$ with the smallest number
of variables from $\vec{A}$ assigned $1$. That is,
MINQSAT$_2(\Phi) = |\vec{W}| - 1$.
}

Now suppose that if $\Phi \in {\rm TQBF}$ and let $v$ be a witness 
for $\Phi$.  Let $\vec{W}$ consist of $C$ together with the variables $S \in
\vec{A}$ such that $v(S)$ is true.
Let $\vec{w}$ set all the variables in $\vec{W}$ to 1.
It is easy to see $X = 0$ is a cause of $\psi$, using this choice of
$\vec{W}$ and $\vec{w}$ for AC2.
Clearly $(M,u) \models [X \gets 1, \vec{W} \gets \vec{1}]\neg\psi$, since
setting $X$ to $1$ and and all variables in $\vec{W}$ to 1
causes all variables in $\vec{B}$ to have the value $2$.
On the other hand, $(M,u) \models [X \gets 0, \vec{W} \gets
\vec{1}, \vec{Z}' \gets 0]\psi$, since setting $X$ to 
0 and $\vec{W} \leftarrow \vec{w}$ guarantees that all variables in
$\vec{B}$  have the value $1$ and $\varphi'$ is satisfied.
It now follows that a minimal set $\vec{W}$ satisfying AC2 consists of
$C$ together with a minimal set of variables in $\vec{A}$ that are true in
a witness for $\Phi$.  Thus, $|\vec{W}| =  {\rm MINQSAT}_2(\Phi) + 1$, 
and we have
$dr((M,\vec{u}),(X=0),\psi) = 1/(|\vec{W}| + 1) = 1/({\rm
MINQSAT}_2(\Phi)+2)$. 
\end{proof}

\subsection{The complexity of computing blame}

Given an epistemic state $(\K,\Pr)$, where $\K$ consists of $N$ possible 
augmented
situations,
each with at most $n$ random variables, the straightforward way
to compute  
$\db(\K,\Pr,X \gets x, \phi)$ is by computing 
$dr((M_{X \gets x, \vec{Y} \gets \vec{y}},\vec{u}),X=x,\varphi)$
for each $(M,\vec{u}, \vec{Y} \gets \vec{y}) \in \K$ 
such that $(M,\vec{u}) \models X \ne x \land \neg \phi$,
and then computing the expected
degree of responsibility 
with respect to these situations, as in Definition~\ref{def:blame}.
Recall that the degree of responsibility in each model 
$M$
is determined by
a binary search, 
and uses at most $\lceil \log{n_M} \rceil$ queries to the oracle, where
$n_M$ is the number of endogenous variables in $M$.
Since there are $N$ augmented situations in $\K$, we get a 
polynomial time algorithm with 
$\sum_{i = 1}^ N \lceil \log{n_i} \rceil$ oracle queries.
Thus, it is clear that the number of queries is at most the size of the
input, and 
is
also at most $N \lceil \log{n^*} \rceil$, where $n^*$ is the
maximum number of endogenous variables  that appear in any of the $N$
augmented situations in $\K$.  
The type of oracle
depends on whether the models are binary or general. For binary models it
is enough to have an NP oracle, whereas for general models we need a
$\Sigma^P_2$-oracle.  
It follows from the discussion above that the problem of computing
the degree of blame in general models \fpsigman, where $n$ is the size
of the input.  However, the best lower bound we can prove is \fpsigma,
by reducing the problem of computing responsibility to that of computing
blame; indeed, the degree of responsibility 
can be viewed as a special case of the degree of blame with the
epistemic state consisting of only one situation.
Similarly, lower and upper bounds of \fp\ and \fpn\ hold for binary models.

An alternative characterization of the complexity of computing blame can
be given by considering the
complexity classes \fpsigmapars  and \fppar, which consist of all
functions that can be computed in polynomial time with
parallel (i.e., non-adaptive) queries to 
a
$\Sigma^P_2$ (respectively,
NP) oracle. (For background on these complexity classes see 
\cite{JT95,Joh90}.)  
Using \fpsigmapars and \fppar, we can get matching upper and lower bounds.

\begin{theorem}\label{theor-complexity-blame}
The problem of computing blame in recursive causal models is
\fpsigmapar-complete.  
The problem is \fppar-complete in binary causal models.
\end{theorem}
\begin{proof}
As we have observed,
the naive algorithm for computing 
the degree of blame uses $N\log{n^*}$ queries,
where $N$ is the number of augmented situations in the epistemic state,
and $n^*$ 
is the maximum number of variables in each one.
However, the answers to oracle queries for 
one situation do not affect the choice of queries for other situations,
thus queries for different situations can be asked in parallel. The
queries for one 
situation are adaptive, however, as shown in \cite{JT95}, a logarithmic
number of adaptive 
queries can be replaced with a polynomial number of non-adaptive queries
by asking 
all possible combinations of queries in parallel. 
Thus, the problem of computing blame is in \fpsigmapar\ for arbitrary
recursive models, and in \fppar\ for binary causal models.

For hardness in the case of arbitrary recursive models, we provide a reduction
from the following 
\fpsigmapar-complete problem 
\cite{JT95,EL02}. 
Given $N$ closed QBFs $\Phi_1, \ldots, \Phi_N$, where 
$\Phi_i = \exists \vec{A}_i \forall \vec{B}_i
\varphi_i$ and $\varphi_i$ is a propositional formula for $1 \leq i \leq N$
of size $O(n)$,
compute the vector $(v_1, \ldots, v_N) \in \{ 0,1 \}^N$ such that
for all $i \in \{ 1, \ldots, N \}$, we have that $v_i = 1$ iff 
$\Phi_i \in TQBF$.
Without loss of generality, we assume that all $\vec{A}_i$'s and 
$\vec{B}_i$'s are disjoint.
We construct an epistemic state $(\K,\Pr)$, a formula $\phi$, and 
a variable $X$ 
such that $(v_1, \ldots, v_N)$ can be computed 
in polynomial time from the 
degree of blame of setting $X$ to 1 for $\phi$ relative to $(\K,\Pr)$. 
$\K$ consists of the $N+1$ augmented situations $(M_i,\vec{u}_i,
\emptyset)$, $i = 1, 
\ldots, N+1$.
Each of the models $M_1, \ldots, M_{N+1}$ 
is of size $O(N n)$ and
involves all the variables
that appear in the causal models constructed in the hardness part of the
proof of Theorem~\ref{theor-resp-general}, together with fresh random
variables $num_1, \ldots, num_{N+1}$.  
For $1 \leq i \leq N$, the equations for the variables in $\Phi_i$ in
the situation $(M_i,\vec{u}_i)$ at time $1$ are the same as in the proof
of Theorem~\ref{theor-resp-general}.  Thus, $\Phi_i \in {\rm TQBF}$ 
iff $X=1$ is a cause of $\varphi_i$ in
$(M_i,\vec{u}_i)$ at time $1$.   In addition, for $i = 1, \ldots, n$,
in $(M_i,\vec{u}_i)$ at time 1, the equations are such that $num_i$ and
$num_{N+1}$ are set to 1 and all other variables (i.e., $num_{j}$ for $j
\notin \{i,N_1\}$ and all  the variables in $\Phi_j$, $j \ne i$)
are set to 0; at time 0, the equations are such that all
variables are set to 0.  The situation $(M_{N+1},\vec{u}_{N+1})$
is such that all variables are set to 0 at both times 0 and 1.
Let $\Pr(M_i, \vec{u}_i) = 1/2^{i(\lceil \log{n}\rceil)}$ for $1 \leq i
\leq N$, 
and let $\Pr(M_{N+1},\vec{u}_{N+1}) =  1 - \Sigma_{i=1}^N 1/2^{i(\lceil
\log{n} \rceil)}$.
Finally, 
let $\phi = \bigwedge_{i=1}^N (num_i \rightarrow \varphi_i) \land num_{N+1}.$ 

Clearly $\phi$ is 
$0$
in all the $N+1$ situations at time 0 (since
$num_{N+1}$ is false at time 0 in all these situations).  At time 1,
$\phi$ is true in $(M_i,\vec{u}_i)$, $i = 1, \ldots, N$, iff $X=1$ is a
cause of $\phi_i$, and $\phi$ is false in $(M_{N+1},\vec{u}_{N+1})$. 
Thus, the degree of blame
$db(\K,\Pr,X \leftarrow x, \phi)$ is 
an $N\lceil \log{n} \rceil$-bit
fraction, where the $i$th group of bits 
of size $n$
is not $0$ iff $\Phi_i \in {\rm TQBF}$.  It immediately follows that
the vector $(v_1, \ldots, v_N)$ can be extracted from $db(\K,\Pr,X
\leftarrow x, \phi)$ 
by assigning $v_i$ the value $1$ iff 
the bits in the $i$th group in $db(\K,\Pr,X \leftarrow x, \phi)$ are not
all $0$.

We can similarly prove that computing degree of blame in binary models
is \fppar-hard
by reduction from the problem of determining which of $N$ given
propositional formulas of size $O(n)$ are satisfiable.  
We leave details to the reader.
\end{proof}

\stam{
We do not have a matching lower bound but, as we show in the full paper,
any binary-search style algorithm 
for computing the degree of blame requires $\Omega(N \log{n})$
oracle queries,
for $N = o(n)$.   
We do not have a matching lower bound, but we are able to prove that 
there exist situations for which the number of possible values for blame is $n^N$, 
and thus any binary-search
style algorithm for computing the degree of blame requires
$\Omega(\log{n^N}) = \Omega(N\log{n})$ queries. We prove the following lemma.
\begin{lemma}
\end{lemma}
\begin{proof}

\end{proof}
} %

\section{Discussion}\label{sec:discussion}
We have introduced definition of responsibility and blame, based on
Halpern and Pearl's definition of causality.  We cannot say that a
definition is ``right'' or ``wrong'', but we can and should examine the
how useful a definition is, particularly the extent to which it captures
our intuitions.

There has been extensive discussion of causality in the
philosophy literature, and many examples demonstrating the subtlety of
the notion.  (These examples have mainly been formulated to show
problems with various definitions of causality that have been proposed
in the literature.)  Thus, one useful strategy for arguing that a
definition of causality is useful is to show that it can handle these
examples well.  This was in fact done for Halpern and Pearl's definition
of causality (see \cite{HP01b}). 
While we were not able to
find a corresponding body of examples  for responsibility and blame in the
philosophy literature, there is a large body of examples in the legal
literature (see, for example, \cite{HH85}).  We plan to do a more
careful analysis of how our framework can be used for legal reasoning in
future work.   For now, we just briefly discuss some relevant issues.
While we believe that ``responsibility'' and ``blame'' as
we have defined them are important, distinct notions,
the words ``responsibility'' and ``blame'' are often used
interchangeably in natural language.  It is not always obvious which
notion is more appropriate in any given situation.
For example, Shafer~\citeyear{Shafer01} says that ``a child who pokes at
a gun's trigger out of curiosity will not be held culpable for resulting
injury or death''.  Suppose that a child does in fact poke a gun's
trigger and, as a result, his father dies.  According to our definition,
the child certainly is a cause of his father's death (his father's not
leaving the safety catch on might also be a cause, of course), and has
degree of responsibility 1.  However, under reasonable assumptions about
his epistemic state, the child might well have degree of blame 0.
So, although we would say that the child is responsible for his father's
death, he is not to blame.
Shafer talks about the need to take intention into account when assessing
culpability.  In our definition, we 
take intention into account to some extent
by considering the agent's epistemic state.
For example, if the child did not consider it possible that pulling the
trigger would result in his father's death, then surely he had no
intention of causing his father's death.  However, to really capture
intention, we need a more detailed modeled of motivation and
preferences.  

Shafer \citeyear{Shafer01} also discusses the probability of assessing
responsibility in cases such as the following.
\begin{example}\label{xam:insecticide}  Suppose Joe sprays
insecticide on his corn field.  It is known that spraying insecticide
increases the probability of catching a cold from 20\% to 30\%.  The
cost of a cold in terms of pain and lost productivity is \$300.  Suppose
that Joe sprays insecticide and, as a result, Glenn catches a cold.
What should Joe pay?

Implicit in the story is a causal model of catching a cold, which is
something like the following.\footnote{This is not the only reasonable
causal model that could correspond to the story, but it is good enough
for our purposes.}
There are 
four variables:
\begin{itemize}
\item a random variable $C$ (for {\em contact}) such that if $C=1$ is
Glenn is in casual contact with someone who has a cold, and 0 otherwise.
\item a random variable $I$ (for {\em immune}) such that if $I=2$, Glenn
does not catch a cold even if he both comes in contact with a cold sufferer
and lives near a cornfield sprayed with insecticide; if $I=1$, Glenn
does not catch a cold even if comes in contact with a cold sufferer,
provided he does not live near a cornfield sprayed with insecticide; and
if $I=0$, then Glenn catches a cold if he comes in contact with a cold
sufferer, whether or not he lives near a cornfield sprayed with
insecticide; 
\item a random variable $S$ which is 1 if Glenn lives near a 
cornfield sprayed with insecticide and 0 otherwise;
\item a random variable $CC$ which is 1 if Glenn catches a cold and 0
otherwise. 
\end{itemize}
The causal equations are obvious: $CC=1$ iff $C=1$ and either $I = 0$ or
$I=1$ and $S=1$.  The numbers suggest that for 70\% of the population,
$I = 2$, for 10\%, $I = 1$, and for 20\%, $I = 0$.  Suppose that no one
(including Glenn) knows whether $I$ is 0, 1, or 2.  Thus, Glenn's expected
loss from a cold is \$60 if Joe does not spray, and \$90 if he does.
The difference of \$30 is the economic cost to Glenn of Joe spraying
(before we know whether Glenn actually has a cold).
  
As Shafer points out, the law does not allow anyone to sue until there
has been damage.  So consider the situation after Glenn catches a cold.  
Once Glenn catches a cold, it is clear that $I$ must be either 0 or 1. 
Based on the statisical information, $I$ is twice as likely to be 0 as
1.  
This leads
to an obvious epistemic state, where the
causal model where $I=0$ is assigned probability $2/3$ and the causal
model where $I=1$ is assigned probability $1/3$.  
In the latter model,
Joe's spraying is not a cause of Glenn's catching a cold; in the former
it is (and has degree of responsibility 1).
Thus, Joe's degree of blame
for the cold is $1/3$.  This suggests that, once Joe sprays and Glenn
catches a cold, the economic damage is \$100. 
\end{example}
This example also emphasizes the importance of distinguishing between
the epistemic states 
before and after the action is taken, an issue already discussed 
in Section~\ref{sec:blame}. Indeed, 
examining the situation after Glenn caught cold enables us to ascribe
probability $0$ to the situation where Glenn is immune, and thus
increases 
Joe's degree of blame for Glenn's cold.

Example~\ref{xam:insecticide}
is a relatively easy one, since the degree of
responsibility is 1.  Things can quickly get more complicated.  
Indeed, a great deal of legal theory is devoted to issues of
responsibility (the classic reference is \cite{HH85}).   There are a
number of different legal principles that are applied in determining
degree of responsibility; some of these occasionally conflict
(at least, they appear to conflict to a layman!).  
For example, in some cases,
legal practice (at least, American legal
practice) does not really consider degree of responsibility as we have
defined it. 
Consider the rock throwing example, and suppose the bottle belongs to
Ned and is somewhat valuable; in fact, it is worth \$100.
\begin{itemize}
\item Suppose both Suzy and Billy's rock hit the bottle simultaneously,
and all it takes is one rock to shatter the bottle.  Then they are both
responsible for the shattering to degree $1/2$ (and both have degree of
blame $1/2$ if this  model is commonly known).  Should both have to pay
\$50?   What if they bear different degrees of responsibility?

Interestingly, a standard legal principle is also that 
``an individual defendant's responsibility does not decrease just because
another wrongdoer was also an actual and proximate cause of the
injury'' (see \cite{pubtrust}).
That is, our notion of having a degree of responsibility less than one is
considered inappropriate 
in some cases
in standard tort law, as is the notion of
different degrees of responsibility.
Note that if Billy is broke and Suzy can afford \$100, the
doctrine of {\em joint and several liability}, also a standard principle in
American tort law, rules that Ned can recover the full \$100 from Suzy.

\item Suppose that instead it requires two rocks to shatter the bottle.
Should that case be treated any differently?  (Recall that, in this
case, both Suzy and Billy have degree of responsibility 1.)

\item If Suzy's rock hits first and it requires only one rock to shatter
the bottle then, as we have seen, Suzy has degree of
responsibility 0 or $1/2$ (depending on whether we consider only allowable
settings) and Billy has degree of responsibility 0.  
Nevertheless, standard legal practice would probably judge Billy (in
part) responsible.
\end{itemize}

In some cases, it seems that legal doctrine confounds what we have
called cause, blame, and responsibility.  To take just one example from
Hart and Honore\'e \citeyear[p.~74]{HH85}, 
assume that
$A$ throws a lighted
cigarette into the bracken near a forest and a fire starts.  Just as the
fire is about to go out, $B$ deliberately pours oil on the flame.  The
fire spreads and burns down the forest.  Clearly $B$'s action was a
cause of the forest fire.  Was $A$'s action also a cause of the forest
fire?  According to Hart and Honor\'e, he is not, whether or not he
intended to cause the fire; only $B$ was.   In our framework, it depends
on the causal model.  If $B$ would have started a fire anyway, whether
or not $A$'s fire went out, then $A$ is indeed not the cause; if $B$
would not have started a fire had he not seen $A$'s fire, then $A$ is a
cause (as is $B$), although is degree of resonsibility for the fire is
only $1/2$.  Furthermore, $A$'s degree of blame may be quite low.
Our framework lets us make distinctions here that seem to be relevant
for legal reasoning.

While these examples show that legal reasoning treats responsibility and
blame somewhat differently from the way we do, we believe that a formal
analysis of legal reasoning using our definitions would be helpful, both
in terms of clarifying the applicability of our definitions and in terms
of clarifying the basis for various legal judgments.  As we said, we
hope to focus more on legal issues in future work.

Our notion of degree 
of
responsibility focuses on when an action becomes
critical.  Perhaps it may have been better termed a notion of ``degree
of criticality''.  While we believe that it is a useful notion, there
are cases where a more refined notion may be useful.  For example,
consider a voter who voted for Mr.~B in the case of a 1-0 vote and a
voter who voted for Mr.~B in the case of a 100-99 vote.  In both case, that
voter has degree of responsibility 1.  While it is true that, in both cases,
that voter's vote was critical, in the second case, the voter may
believe that his responsibility is more diffuse.  We often here
statements like ``Don't just blame me; I wasn't the only one who voted for
him!''  The second author is current working with I. Gilboa on a
definition of resonsibility that uses the game-theoretic notion of {\em
Shapley value\/} (see, for example, \cite{OR94}) to try to distinguish
these examples.  

As another example, suppose that one person dumps 900
pounds of garbage on a porch and another dumps 100 pounds.  The porch
can only bear 999 pounds of load, so it collapses.  Both people here
have degree of responsibility $1/2$ according to our definition, but
there is an intuition that 
suggests that the person who dumped 900 pounds should bear greater
responsibliity.  We can easily accommodate this in our framework by
putting weights on variables.  If we use $\wt(X)$ to denote the weight of a
 $X$ and $\wt(\vec{W})$ to denote the sum of the weights of variables in
the set $\vec{W}$, then we can define
the degree of responsibility of $X=x$ for $\phi$ to be
$\wt(X)/(\wt(\vec{W}) + \wt(X)$, where $\vec{W}$ is a set of minimal
weight for which AC2  holds.  This definition agrees with the one we use
if the weights of all variables are 1, so this can be viewed as a
generalization of our current definition.   

These examples show that there is much more to be done in clarifying our
understanding of responsibility and blame.  Because these notions are so
central in law and morality, we believe that doing so is quite worthwhile.

\paragraph*{Acknowledgment} We thank Michael Ben-Or
and Orna Kupferman
for helpful discussions.
We particularly thank Chris Hitchcock, who suggested a simplification of
the definition of blame and pointed 
out a number of typos and other problems in an
earlier version of the paper. 

\bibliographystyle{chicago}
\bibliography{z,joe}

\begin{thebibliography}{}

\bibitem[\protect\citeauthoryear{Chockler, Halpern, and Kupferman}{Chockler
  et~al.}{2003}]{CHK}
Chockler, H., J.~Y. Halpern, and O.~Kupferman (2003).
\newblock What causes a system to satisfy a specification?
\newblock Unpublished manuscript. Available at
  http://www.cs.cornell.edu/home/halpern/papers/resp.ps.

\bibitem[\protect\citeauthoryear{Collins, Hall, and Paul}{Collins
  et~al.}{2003}]{Collins03}
Collins, J., N.~Hall, and L.~A. Paul (Eds.) (2003).
\newblock {\em Causation and Counterfactuals}.
\newblock Cambridge, Mass.: MIT Press.

\bibitem[\protect\citeauthoryear{Eiter and Lukasiewicz}{Eiter and
  Lukasiewicz}{2002a}]{EL02}
Eiter, T. and T.~Lukasiewicz (2002a).
\newblock Causes and explanations in the structural-model approach: tractable
  cases.
\newblock In {\em Proc.~Eighteenth Conference on Uncertainty in Artificial
  Intelligence (UAI 2002)}, pp.\  146--153.

\bibitem[\protect\citeauthoryear{Eiter and Lukasiewicz}{Eiter and
  Lukasiewicz}{2002b}]{EL01}
Eiter, T. and T.~Lukasiewicz (2002b).
\newblock Complexity results for structure-based causality.
\newblock {\em Artificial Intelligence\/}~{\em 142\/}(1), 53--89.

\bibitem[\protect\citeauthoryear{Hall}{Hall}{2003}]{Hall98}
Hall, N. (2003).
\newblock Two concepts of causation.
\newblock In J.~Collins, N.~Hall, and L.~A. Paul (Eds.), {\em Causation and
  Counterfactuals}. Cambridge, Mass.: MIT Press.

\bibitem[\protect\citeauthoryear{Halpern and Pearl}{Halpern and
  Pearl}{2001a}]{HP01b}
Halpern, J.~Y. and J.~Pearl (2001a).
\newblock Causes and explanations: A structural-model approach. {P}art {I}:
  {C}auses.
\newblock In {\em Proc.~Seventeenth Conference on Uncertainty in Artificial
  Intelligence (UAI 2001)}, pp.\  194--202.
\newblock The full version of the paper is available at
  http://www.cs.cornell.edu/home/halpern.

\bibitem[\protect\citeauthoryear{Halpern and Pearl}{Halpern and
  Pearl}{2001b}]{HP01a}
Halpern, J.~Y. and J.~Pearl (2001b).
\newblock Causes and explanations: A structural-model approach. {P}art {II}:
  {E}xplanations.
\newblock In {\em Proc.~Seventeenth International Joint Conference on
  Artificial Intelligence (IJCAI '01)}, pp.\  27--34.
\newblock The full version of the paper is available at
  http://www.cs.cornell.edu/home/halpern.

\bibitem[\protect\citeauthoryear{Hart and Honor{\'e}}{Hart and
  Honor{\'e}}{1985}]{HH85}
Hart, H.~L.~A. and T.~Honor{\'e} (1985).
\newblock {\em Causation in the Law\/} (Second Edition ed.).
\newblock Oxford, U.K.: Oxford University Press.

\bibitem[\protect\citeauthoryear{Hume}{Hume}{1739}]{Hume39}
Hume, D. (1739).
\newblock {\em A Treatise of Human Nature}.
\newblock London: John Noon.

\bibitem[\protect\citeauthoryear{Jenner and Toran}{Jenner and
  Toran}{1995}]{JT95}
Jenner, B. and J.~Toran (1995).
\newblock Computing functions with parallel queries to {NP}.
\newblock {\em Theoretical Computer Science\/}~{\em 141}, 175--193.

\bibitem[\protect\citeauthoryear{Johnson}{Johnson}{1990}]{Joh90}
Johnson, D. (1990).
\newblock A catalog of complexity classes.
\newblock In J.~van Leeuwen (Ed.), {\em Handbook of Theoretical Computer
  Science}, Volume~A, Chapter~2. Elsevier Science.

\bibitem[\protect\citeauthoryear{Osborne and Rubinstein}{Osborne and
  Rubinstein}{1994}]{OR94}
Osborne, M.~J. and A.~Rubinstein (1994).
\newblock {\em A Course in Game Theory}.
\newblock Cambridge, Mass.: MIT Press.

\bibitem[\protect\citeauthoryear{Papadimitriou}{Papadimitriou}{1984}]{Pap84}
Papadimitriou, C.~H. (1984).
\newblock The complexity of unique solutions.
\newblock {\em Journal of {ACM}\/}~{\em 31}, 492--500.

\bibitem[\protect\citeauthoryear{Pearl}{Pearl}{2000}]{pearl:2k}
Pearl, J. (2000).
\newblock {\em Causality: Models, Reasoning, and Inference}.
\newblock New York: Cambridge University Press.

\bibitem[\protect\citeauthoryear{{Public Trust}}{{Public
  Trust}}{2003}]{pubtrust}
{Public Trust} (2003).
\newblock Tort law issues/fact sheets.
\newblock
  http://www.citizen.org/congress/civjus/tort/tortlaw/articles.cfm?ID=834).

\bibitem[\protect\citeauthoryear{Shafer}{Shafer}{1996}]{Shafer96}
Shafer, G. (1996).
\newblock {\em The Art of Causal Conjecture}.
\newblock Cambridge, Mass.: MIT Press.

\bibitem[\protect\citeauthoryear{Shafer}{Shafer}{2001}]{Shafer01}
Shafer, G. (2001).
\newblock Causality and responsibility.
\newblock {\em Cardozo Law Review\/}~{\em 22}, 101--123.

\bibitem[\protect\citeauthoryear{Stockmeyer}{Stockmeyer}{1977}]{Stock}
Stockmeyer, L.~J. (1977).
\newblock The polynomial-time hierarchy.
\newblock {\em Theoretical Computer Science\/}~{\em 3}, 1--22.

\bibitem[\protect\citeauthoryear{Zimmerman}{Zimmerman}{1988}]{Zimmerman88}
Zimmerman, M. (1988).
\newblock {\em An Essay on Moral Responsibility}.
\newblock Totowa, N.J.: Rowman and Littlefield.

\end{thebibliography}

\end{document}